# Vision Transformers Exhibit Human-Like Biases: Evidence of Orientation and Color Selectivity, Categorical Perception, and Phase Transitions

Nooshin Bahador

Krembil Research Institute, University Health Network, Toronto, Canada.

**Abstract**

This study explored whether Vision Transformers (ViTs) developed orientation and color biases similar to those observed in the human brain. Using synthetic datasets with controlled variations in noise levels, angles, lengths, widths, and colors, we analyzed the behavior of ViTs fine-tuned with LoRA. Our findings revealed four key insights: First, ViTs exhibited an "oblique effect" showing the lowest angle prediction errors at 180° (horizontal) across all conditions. Second, angle prediction errors varied by color. Errors were highest for bluish hues and lowest for yellowish ones. Additionally, clustering analysis of angle prediction errors showed that ViTs grouped colors in a way that aligned with human perceptual categories. In addition to orientation and color biases, we observed phase transition phenomena. While two phase transitions occurred consistently across all conditions, the training loss curves exhibited delayed transitions when color was incorporated as an additional data attribute. Finally, we observed that attention heads in certain layers inherently develop specialized capabilities, functioning as task-agnostic feature extractors regardless of the downstream task. These observations suggest that biases and properties arise primarily from pre-training on the original dataset—which shapes the model's foundational representations—and the inherent architectural constraints of the vision transformer, rather than being solely determined by downstream data statistics.

## 1. Introduction

The brain's ability to selectively process specific visual features, such as orientation, is a fundamental aspect of sensory perception. Orientation selectivity arises from the tuning of neurons in the early visual cortex, where cells respond preferentially to stimuli aligned along particular angles (e.g., horizontal or vertical orientations) (Duong et al., 2023). These orientation-tuned cells exhibit distinct response profiles when presented with oriented grating stimuli, with some neurons showing maximal activity for horizontal orientations while others prefer different angles (Duong et al., 2023). This selectivity emerges from a combination of feedforward inputs and recurrent interactions within neural populations, where excitatory and inhibitory inputs from similarly tuned neurons shape the overall tuning properties (Duong et al., 2023). The mechanistic recurrent population model explains how biased orientation probability in natural scenes influences the joint coordination among neurons, leading to anisotropic processing—where certain orientations (e.g., horizontal and vertical) are more efficiently detected than oblique ones (Mansfield, 1974).

This anisotropy—direction-dependent sensitivity in visual processing—is further reflected in the contrast sensitivity function (CSF), which measures how well different spatial frequencies and orientations are detected (Akbarinia et al., 2023). The CSF reveals that the visual system is optimized for processing cardinal (horizontal and vertical) orientations, a phenomenon linked to the statistical regularities of natural environments (Akbarinia et al., 2023). Additionally, meridian asymmetry highlights differences in processing along the horizontal versus vertical axes, with behavioral and neural evidence showing enhanced sensitivity for horizontal orientations (Akbarinia et al., 2023). This asymmetry may arise from the structural anisotropy of natural scenes, where features often extend more prominently along horizontal or vertical axes (Zheng et al., 2023).

A related phenomenon is categorical color perception, where the brain groups colors into distinct categories (e.g., red vs. green) rather than processing them along a continuous spectrum. Just as orientation selectivity reflects tuning to specific angles, color categorization emerges from neural mechanisms that prioritize certain hues over others, likely due to their ecological relevance (Himmelberg et al., 2025). Studies also report higher detection rates for color targets compared to orientation or size, suggesting that color may engage specialized processing pathways (Himmelberg et al., 2025).

The development of these perceptual biases—whether in orientation or color—is shaped by experience and learning. Both humans and artificial neural networks exhibit non-homogeneous sensitivity patterns, where frequently encountered features (e.g., cardinal orientations) are learned earlier and processed more efficiently (Benjamin et al., 2022). This aligns with the oblique effect, where sensitivity to near-vertical and horizontal orientations surpasses that of oblique angles, mirroring the statistical distribution of orientations in natural scenes. Deep networks trained on natural images replicate these perceptual biases, further supporting the idea that learning dynamics drive the emergence of non-uniform sensitivity (Aribenjamin et al., 2022).

Thus, the brain's selective processing of orientation and color reflects an interplay between neural tuning, environmental statistics, and developmental learning—a framework that extends to other perceptual domains, including shape orientation and object-based attention (Blazek et al., 2024).

Since the responses of visual cortical cells are known to depend strongly on stimulus properties such as shape, position, and orientation (Hubel & Wiesel, 1962), we sought to investigate whether Vision Transformers (ViTs) (Dosovitskiy et al., 2020) exhibit similar orientation and color selectivity biases. To test this hypothesis, we designed controlled experiments using synthetically generated training data, allowing us to systematically probe these potential biases in ViT architectures. We also sought to investigate whether certain emergent properties observed in transformer-based language models—such as the formation of induction heads—might also manifest in vision transformers. Induction heads are specialized attention circuits that facilitate in-context learning by detecting and completing repeating patterns, a process often accompanied by a distinct phase change in the training loss curve (marked by a noticeable bump) (Olsson et al., 2022).

## 2. Methods

To investigate whether Vision Transformers (ViTs) exhibit (1) orientation selectivity (neural preference for stimulus orientation), (2) categorical color perception (discrete grouping of color spectra), (3) color selectivity (preferential responses to specific color ranges), and (4) phase transitions (shifts in network dynamics), we generated four synthetic datasets with systematically varied line geometry, color, and noise. We then fine-tuned a Vision Transformer using these progressively complex datasets, employing LoRA-adapted attention layers and multi-task regression heads to predict multiple line properties under controlled conditions.

### 2.1. Controlled Synthetic Datasets

Four progressively complex synthetic datasets ($varying\ angles \rightarrow lengths \rightarrow widths \rightarrow colors$) were generated to systematically test how geometric and chromatic line properties affect fine-tuned vision transformer ($ViT$) performance under controlled noise conditions. ***Appendix I*** explains the mathematical steps involved in generating each of the four synthetic datasets. Figure 1 illustrates the progressive enhancement of line properties across four synthetic datasets (shown through sample generated images).

(A) White Lines with Varied Angles
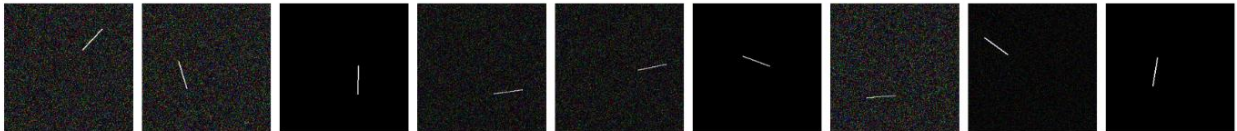

(B) White Lines with Varied Angles, Lengths
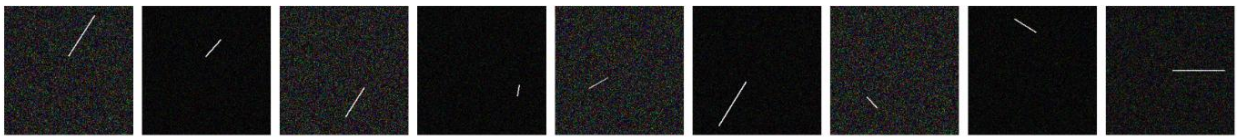

(C) White Lines with Varied Angles, Lengths, and Widths
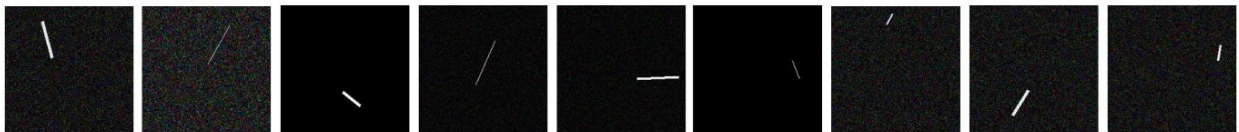

(D) Lines with Varied Angles, Lengths, Widths, and Colors
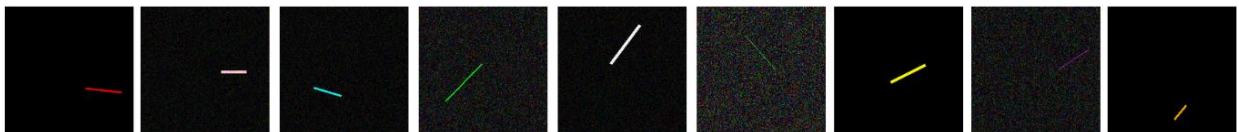

**Fig.1.** Progressive complexity in synthetic datasets: (A) angle variation, (B) added length variation, (C) added width variation, and (D) added color variation.

## 2.2. Fine-Tuning Process of Pre-Trained $ViT$ Model

Figure 2 illustrates the model's key components: a frozen Vision Transformer (ViT) for feature extraction, trainable LoRA-adapted attention layers for efficient fine-tuning, and specialized regression heads that predict multiple line properties in parallel.

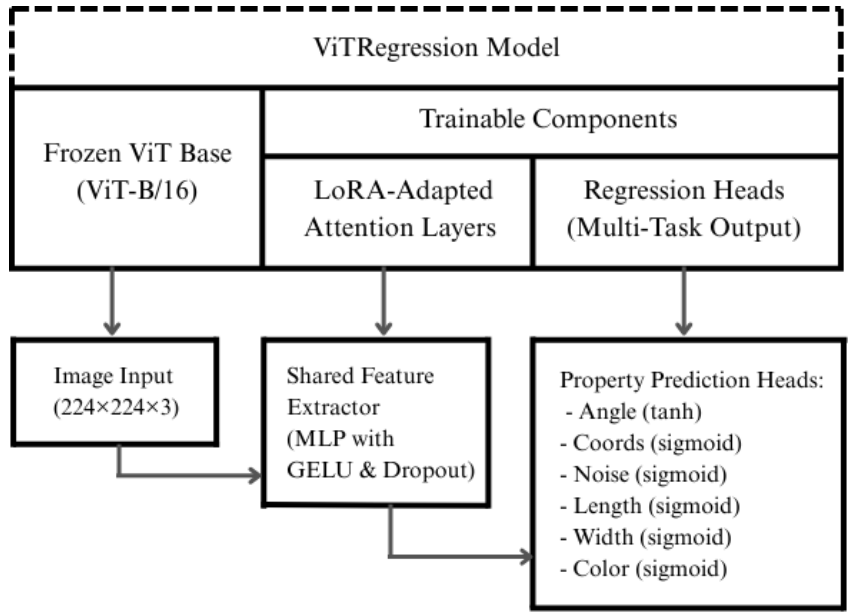

**Fig.2.** Architecture diagram of the ViTRegression model showing frozen ViT backbone, trainable LoRA layers, and multi-task regression heads.

Table 1 summarizes the key components of the fine-tuning protocol, where a pre-trained $ViT$ model is adapted using $LoRA$ for multi-task regression on synthetic line images.

**Table 1**: Vision Transformer Fine-Tuning Protocol Summary

| Component | Specification | Mathematical Formulation |
|---|---|---|
| Dataset | $50k$ synthetic images ($224 \times 224$) with annotated line properties | $I \in \mathbb{R}^{3 \times 224 \times 224}$, $\hat{y} = \{\theta \in [-1,1], p_{start}, p_{pend} \in [0,1]^2, \dots\}$ |
|  | Data Normalization | • Angles: $\theta_{norm} = \frac{\theta_{deg} \bmod 360}{180} - 1$ <br> • Coordinates: $p = p_{px}/224$ <br> • Colors: $c = c_{RGB}/255$ |
| Model | $ViT$-Base + $LoRA$ ($r = 8$) + Multi-task heads | $f_\theta(x) = \{h_{task}(g\phi(x))\}$ <br> *where $\phi$ frozen, LoRA on attention* |
|  | Architecture | • Shared feature extractor: $g_\phi: \mathbb{R}^{3 \times 224 \times 224} \to \mathbb{R}^{768}$ <br> • Task heads: $h_{task}: \mathbb{R}^{768} \to \mathbb{R}^{d_{task}}$ <br> *with appropriate output activations* |
| Training | AdamW ($lr = 1e - 4$), mixed precision, Huber loss | $\mathcal{L} = \sum_{task} w_{task} \mathcal{L}_{Huber}(\hat{y}, y)$ |
| | Loss Weights | • $Angle$: 2.0, $Coordinates$: 1.0, $Other\ properties$: 0.5 |
| Optimization | Reduce LR On Plateau ($patience = 3$), early stopping ($patience = 5$) | $lr_{t+1} = \gamma lr_t\ if\ \mathcal{L}_{val}\ plateaus$ |
| Evaluation | Correlation coefficients and error distributions per property | $\rho_{task} = corr(\hat{y}_{task}, y_{task})$ |

Table 2 summarizes the key characteristics of $ViT$-based regression models for each dataset, highlighting the progressive addition of predicted line properties (angle, length, width, and color) across experiments.

**Table 2:** Comparison of Line Detection Datasets and Fine-Tuned Models

| Component | Dataset I | Dataset II | Dataset III | Dataset IV |
|---|---|---|---|---|
| Target feature for prediction | Angle, start/end points, noise | Angle, start/end points, noise, length | Angle, start/end points, noise, length, width | Angle, start/end points, noise, length, width, color (RGB) |
| Model Output Heads | Angle, coords (4), noise | Angle, coords (4), noise, length | Angle, coords (4), noise, length, width | Angle, coords (4), noise, length, width, color (3) |
| Loss Function | Weighted Huber loss (angle×2, coords×1, others×0.5) | Same as I with length added | Same as II with width added | Same as III with color added |
| Evaluation Metrics | Loss curves, feature correlations, error distributions | Same as I with length metrics added | Same as II with width metrics added | Same as III with color channel added metrics |

## 3. Results

Based on the results obtained from this study and comparison to our previous work (Bahador, 2025), head 4, head 5, and head 12 in layer II of the ViT-Base model, fine-tuned with LoRA, exhibit consistent edge-detection specialization (monosemantic attention heads) across two distinct tasks—chirp localization in 2D spectrograms and line property prediction (Figure 3). This implies that these heads may inherently develop edge-detection capabilities regardless of the downstream task, highlighting their role as task-agnostic feature extractors. The recurrence of this behavior in different fine-tuning scenarios indicates that certain attention heads in vision transformers may stabilize into specific, interpretable functions, such as edge detection, which could be a generalizable property of the model's architecture rather than a task-specific adaptation. This consistency across tasks raises questions about the intrinsic specialization of attention heads and their potential reuse or preservation during fine-tuning.

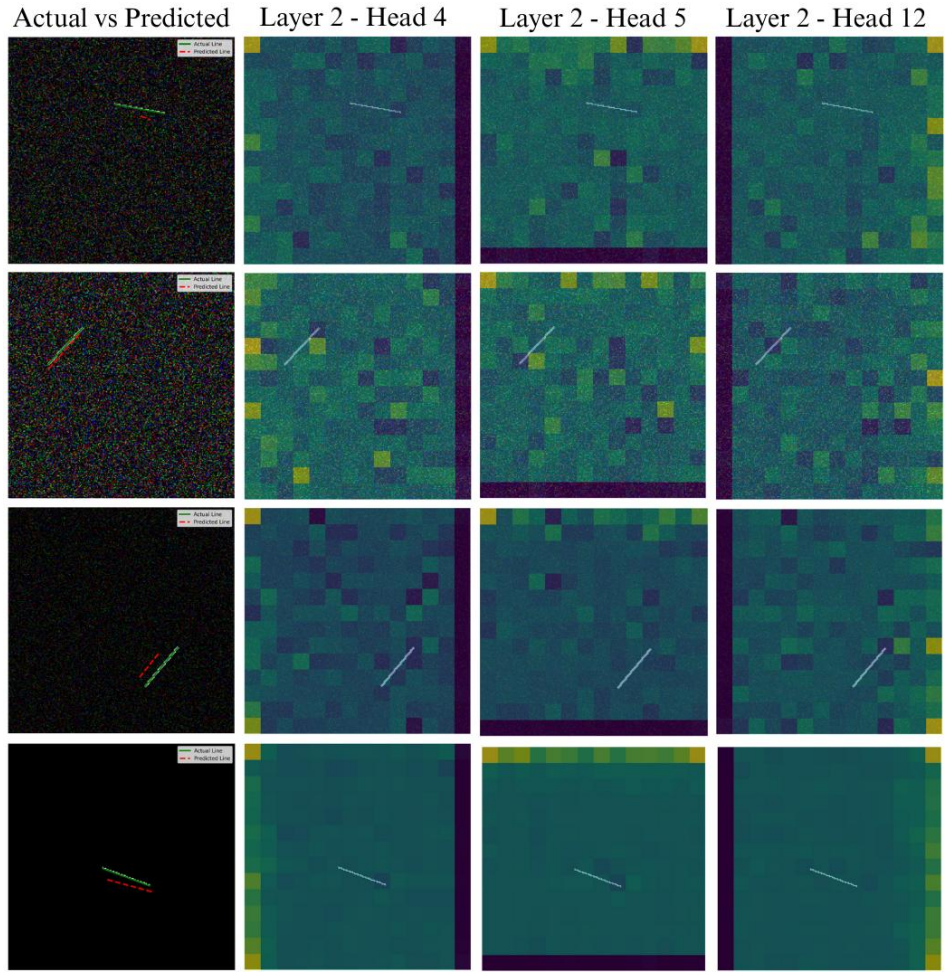

**Fig.3.** Heads 4, 5, and 12 in Layer II of the ViT-Base model (fine-tuned with LoRA on Dataset I) exhibit consistent edge-detection specialization

Figure 4 compares the training loss, test loss, and inference time patterns for models trained on progressively complex datasets (from $angle-$only to $angle + length + width + color$ variants), revealing how added complexity affects learning. The presence of two distinct phase transitions (bumps) in the training loss curves (Figure 4) suggests that the model undergoes structured, stage-like learning across different datasets. These bumps likely mark critical developmental milestones where the model consolidates key computational subcircuits necessary for task performance. The timing of these capabilities' emergence varies depending on the dataset, suggesting a link between data characteristics and when specific circuit develop. Dataset I exhibits the earliest phase transitions, implying rapid formation of critical circuits. In contrast, Datasets II and III demonstrate a delayed first bump, indicating that the model requires more exposure to the data before reaching the initial phase transition. Dataset IV, with the latest phase transitions, points to an even higher abstraction threshold, where the model must accumulate substantial evidence before forming the necessary representations. The sequential nature of these bumps—where the second transition consistently follows the first—implies a hierarchical learning process, with later capabilities building upon earlier ones. Since these transitions correlate with in-context learning, earlier bumps (as in Dataset I) indicate quicker development of this ability, while later bumps (Dataset IV) suggest a more prolonged acquisition period. Timing variations across datasets suggest that phase transitions—an inherent emergent property of vision transformers—are modulated by data properties. Thus, earlier transitions signify more easily discoverable structures, whereas delayed transitions reflect greater learning challenges, requiring extended training to overcome.

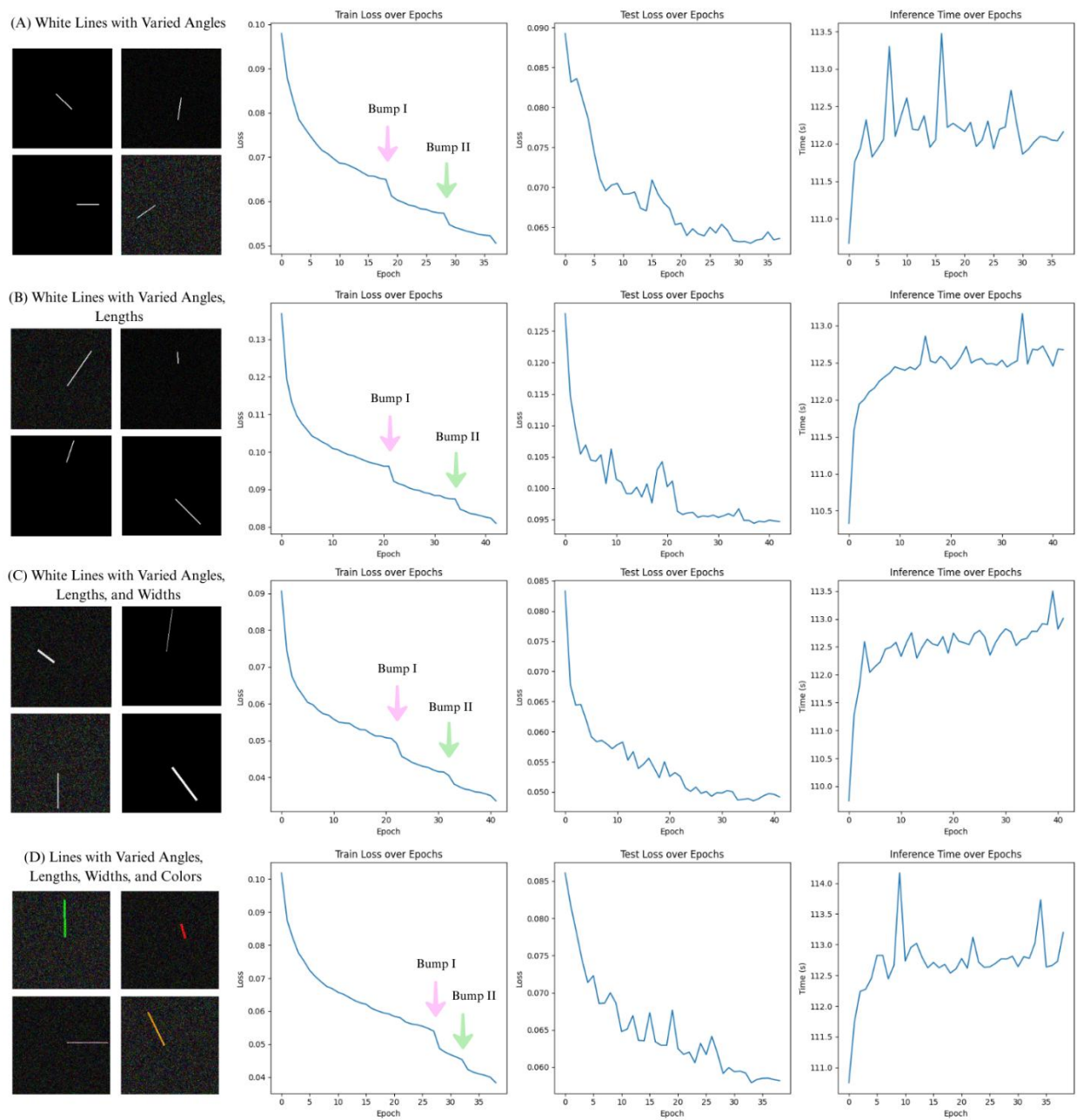

**Fig.4.** Training dynamics across four line-detection datasets showing loss curves and inference times for different line property combinations.

The identical phase transitions observed in the training loss curves consistently reappear in the correlation dynamics of predicted features across training epochs (Figure 5), providing evidence that these bumps reflect genuine developmental milestones rather than incidental fluctuations in optimization. Their synchronized emergence across distinct metrics confirms that these transitions mark critical stages of circuit formation, with timing dictated by dataset-specific learning demands.

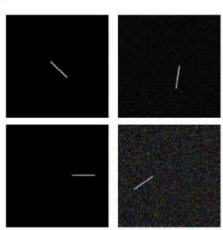
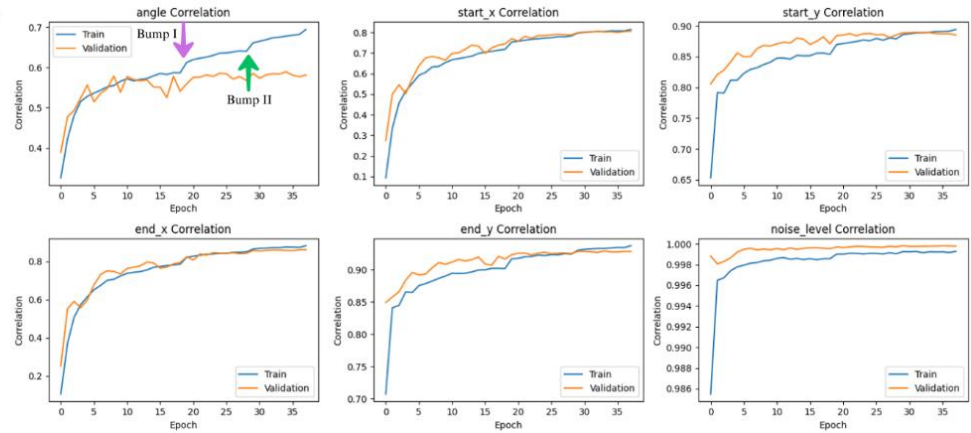

(A) White Lines with Varied Angles

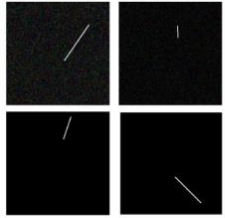
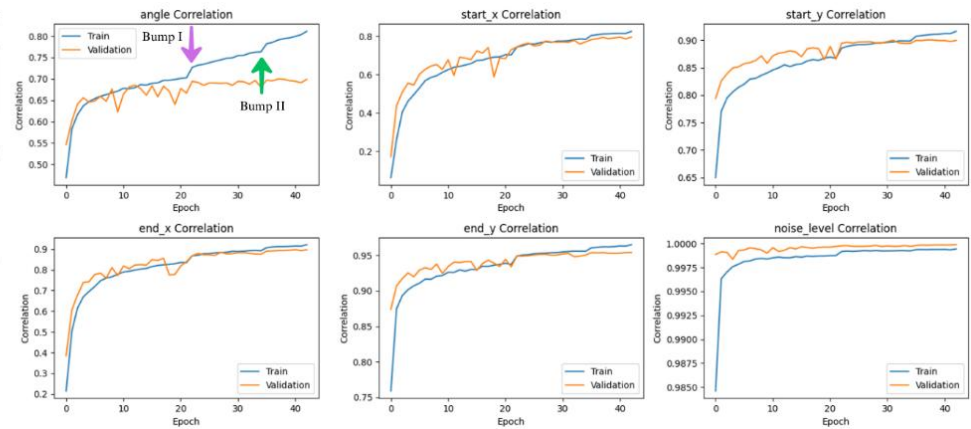

(B) White Lines with Varied Angles, Lengths

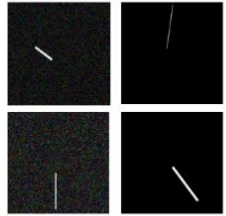
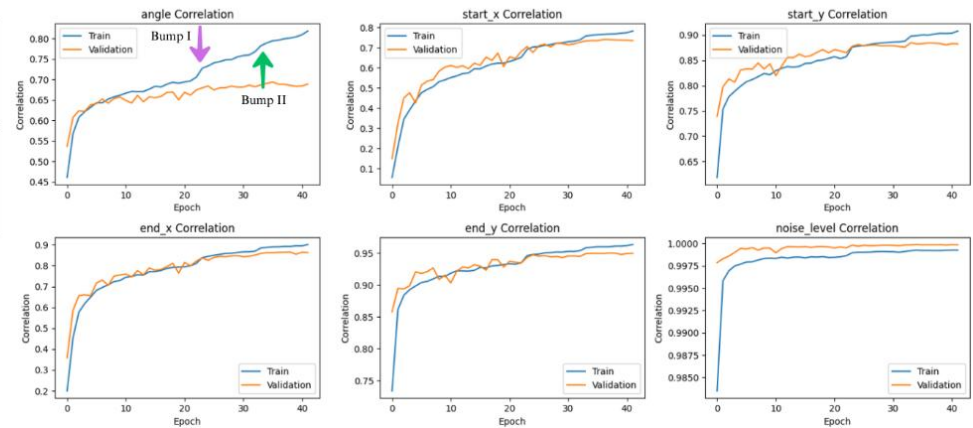

(C) White Lines with Varied Angles, Lengths, and Widths

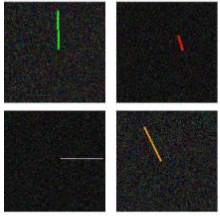
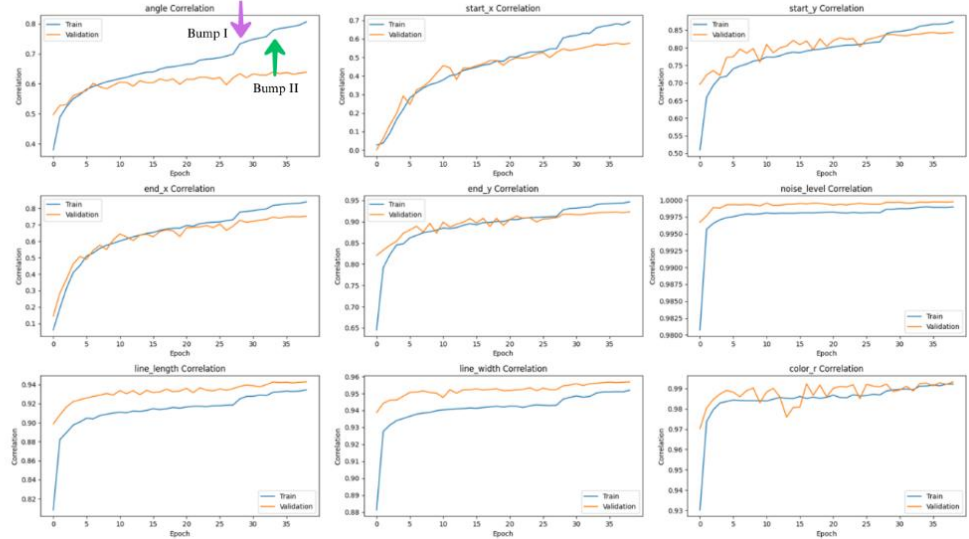

**Fig.5.** Phase transitions in training loss align with feature correlation dynamics, indicating dataset-dependent developmental milestones in circuit formation.

Figure 6 visualizes how prediction error varies with ground-truth line angles, revealing a consistent performance advantage for horizontal orientations (180°) across all tested conditions (length, width, color). The observation that the minimum difference between predicted and ground-truth angles consistently occurs at 180° (horizontal orientation) across all datasets—independent of line size or color—suggests a bias toward horizontal orientations in the model's performance. This phenomenon aligns with neural and computational principles in human brain visual systems. In the primary visual cortex (V1), for instance, horizontally tuned neurons are not only more numerous but also exhibit narrower tuning widths and steeper tuning slopes compared to those tuned to other orientations, leading to heightened sensitivity and precision for horizontal stimuli. This anisotropy, known as the "oblique effect" is reflected in human psychophysics (Edwards et al., 1972), where discrimination thresholds are lowest for horizontal orientations. Prioritization of horizontal features seems to be developed for efficient coding. This bias may also emerge from the network's architecture favoring horizontal alignments. Moreover, the invariance to line size and color implies that this bias operates at a fundamental representational level, likely tied to early feature extraction stages where orientation selectivity is most pronounced.

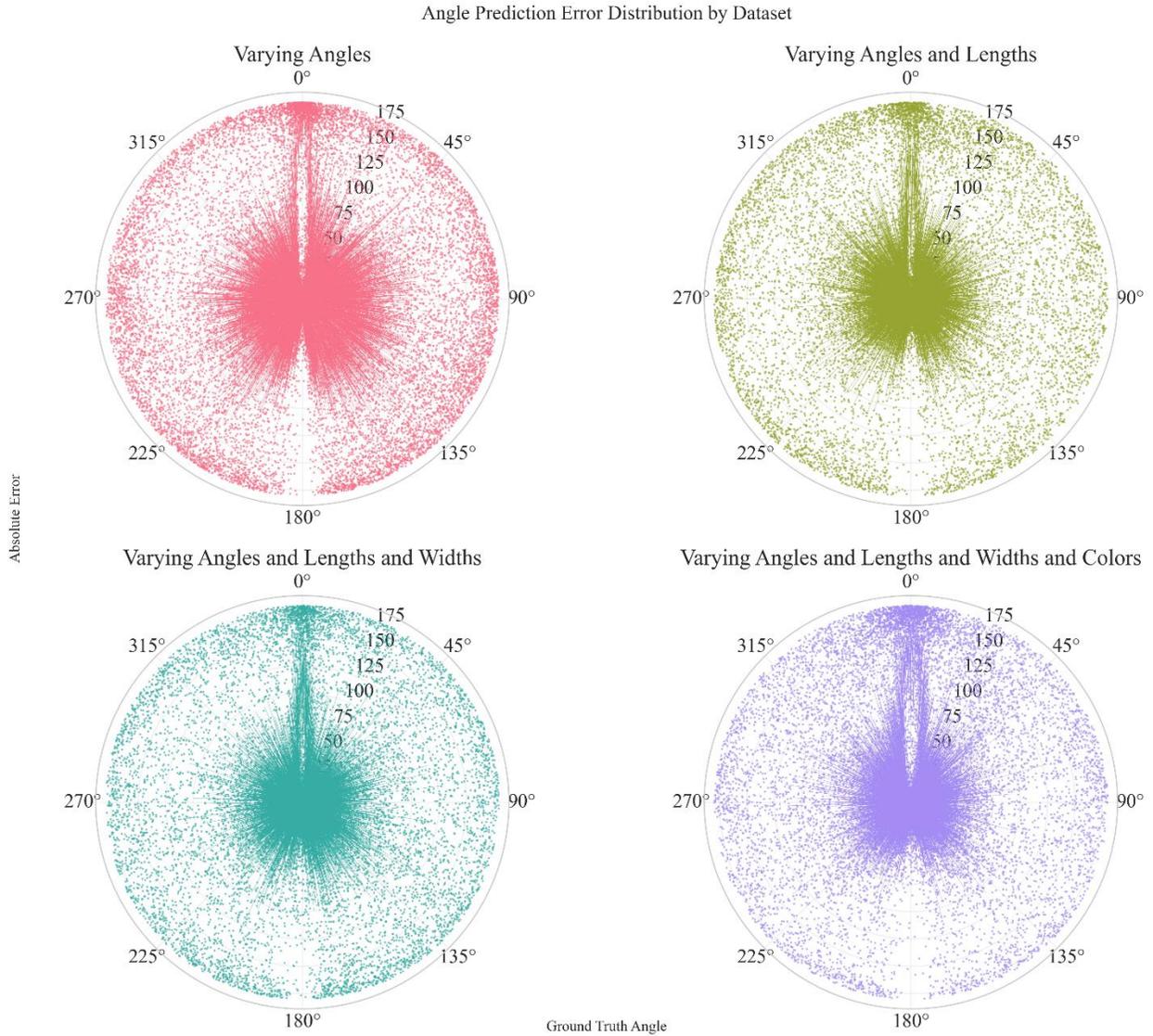

**Fig.6.** Polar plots of angle prediction error across four datasets, showing lowest error at 180° (horizontal) regardless of line variations.

Figure 7 shows how angle prediction errors vary with line length, revealing consistent error patterns across different experimental manipulations of line properties. Median angle errors consistently decrease from ~30° in the shortest bins $(20-28px)$ to ~$1.5-2.5°$ in the longest bins $(92-100px)$ across all datasets ($Dataset\ II$: 30.0 → 1.4; $III$: 30.7 → 1.7; $IV$: 33.5 → 2.4), with largest error reductions typically occurring in mid-length bins.

Hexbin plots (Figure 8) also reveal consistent angle prediction error reduction with longer line lengths across all datasets, though increased variability (widths, colors) in Dataset III–IV introduces slightly higher errors.

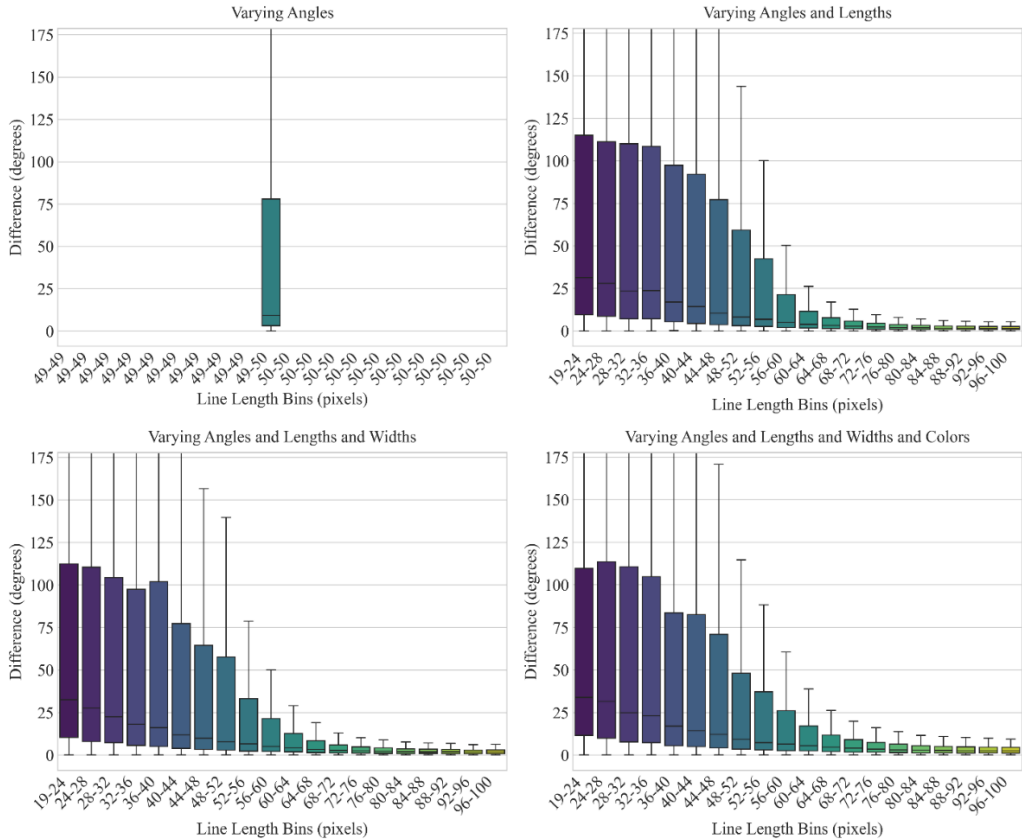

**Fig.7.** Boxplots of angle prediction error distribution across binned line lengths for four experimental conditions. Median error consistently decreases with higher bin ranges: from 30.0 to 1.4 (Dataset II - White Lines with Varied Angles, Lengths), 30.72 to 1.72 (Dataset III - White Lines with Varied Angles, Lengths, and Widths), and 33.46 to 2.41 (Dataset IV - Lines with Varied Angles, Lengths, Widths, and Colors)

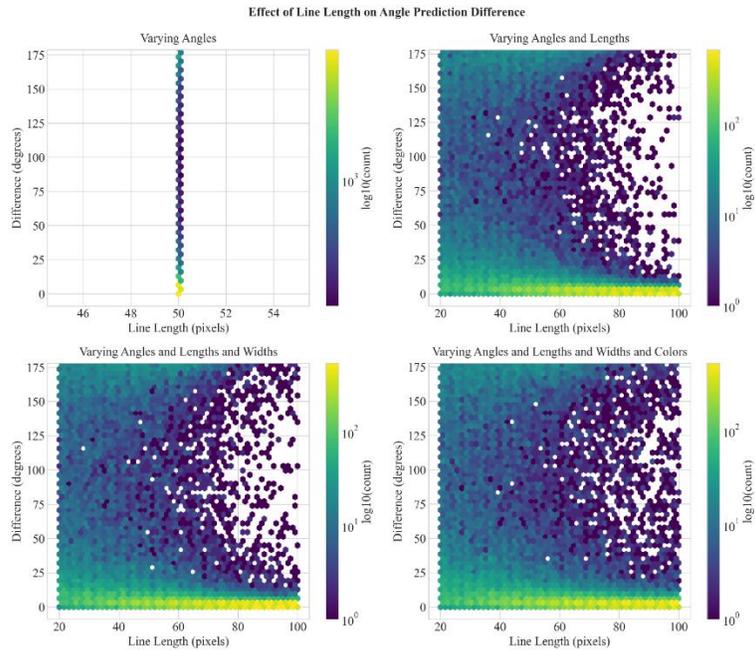

**Fig.8.** Hexbin visualization of angle prediction error versus line length across four datasets

Figure 9 illustrates how the angle prediction difference varies with line width. According to Figure 9, adding color to lines increases angle prediction errors (higher mean differences) compared to monochrome lines, especially for thinner widths (Width 1: 39.49° vs 32.33°).

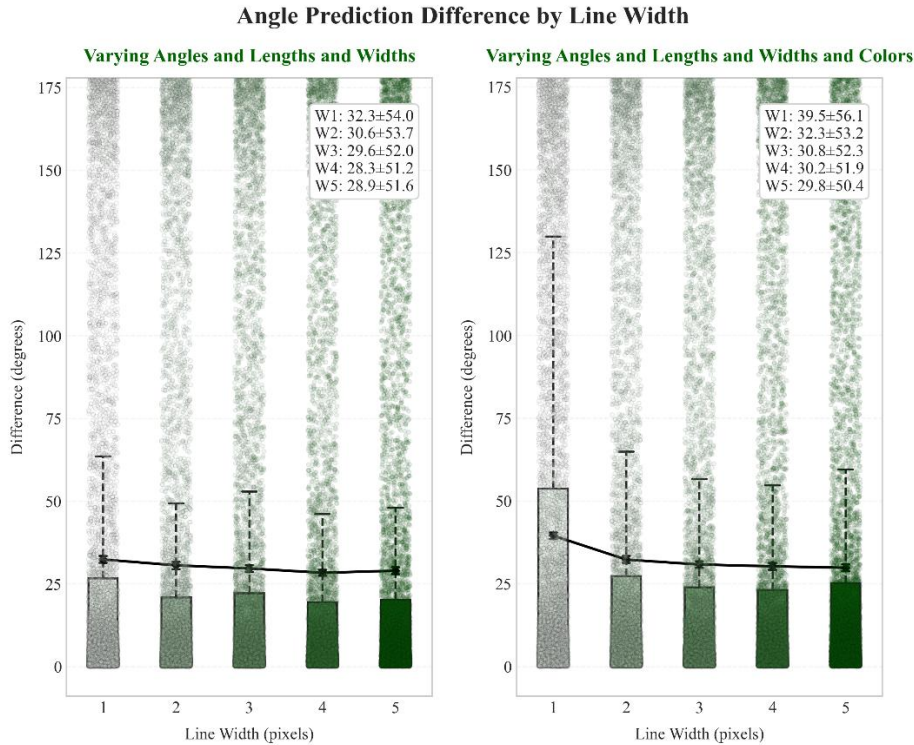

**Fig.9.** Angle prediction error versus line width for monochrome and colored lines: Wider lines exhibit lower errors, while thinner lines (especially Width 1) show higher sensitivity to color

According to Figure 10, the top 3 colors with the largest mean angle differences are blue (mean = 38.12°, median = 7.19°), purple (mean = 37.41°, median = 7.77°), and green (mean = 35.27°, median = 6.16°), while the top 3 colors with the smallest mean angle differences are white (mean = 28.94°, median = 4.81°), pink (mean = 27.32°, median = 5.07°), and orange (mean = 27.00°, median = 5.25°). The colors were clustered into four groups based on their 75th percentile values (Figure 11), using Euclidean distance to minimize within-cluster variance. The 75th percentile represents the value below which 75% of the data falls—indicating that three-quarters of the observations are lower and one-quarter are higher. The resulting color clusters were as follows: (1) Blue, Purple (2) Green, Teal (3) Magenta, Red (4) Cyan, Yellow, White, Pink, Orange. Clustering analysis revealed that bluish hues were associated with the highest errors in angle prediction, while yellowish hues had the lowest. This pattern suggests that Vision Transformers (ViTs), like biological vision systems, exhibit non-uniform sensitivity to color. Just as the human brain processes colors categorically, ViTs may also treat certain color groups differently. The elevated error for blue/purple hues may reflect reduced sensitivity to these wavelengths. These observations support the notion that ViTs develop perceptual biases similar to those in human vision. Moreover, the fact that color groups with similar angle prediction errors also appear perceptually similar to humans further reinforces this connection.

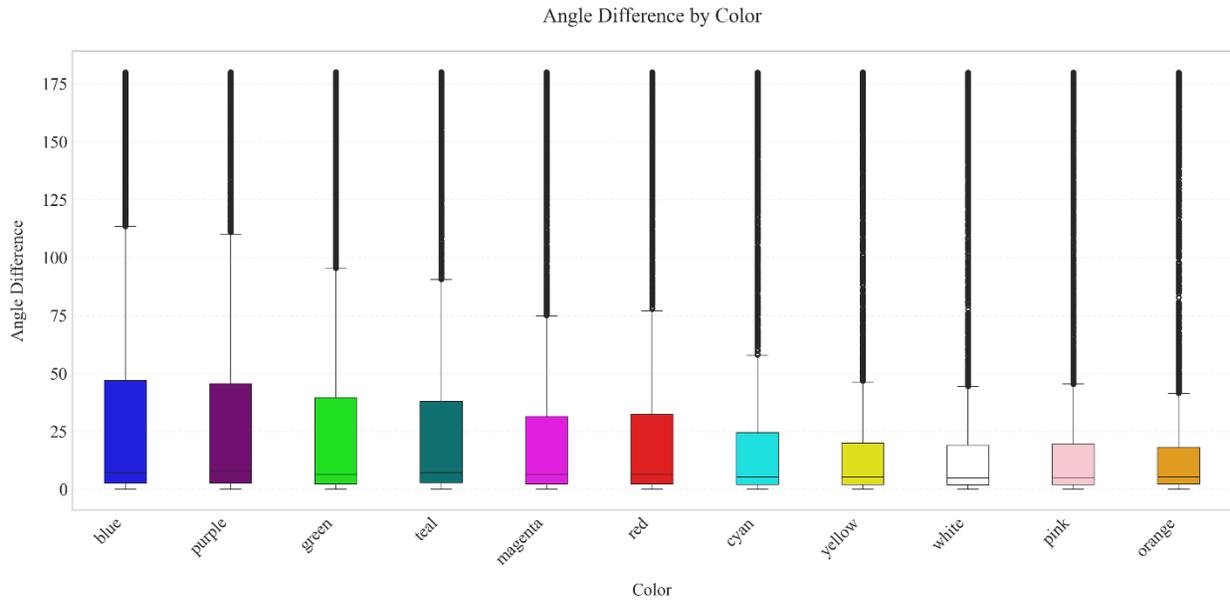

**Fig.10.** Box plot shows the distribution of angle differences across different line colors, ordered by median difference

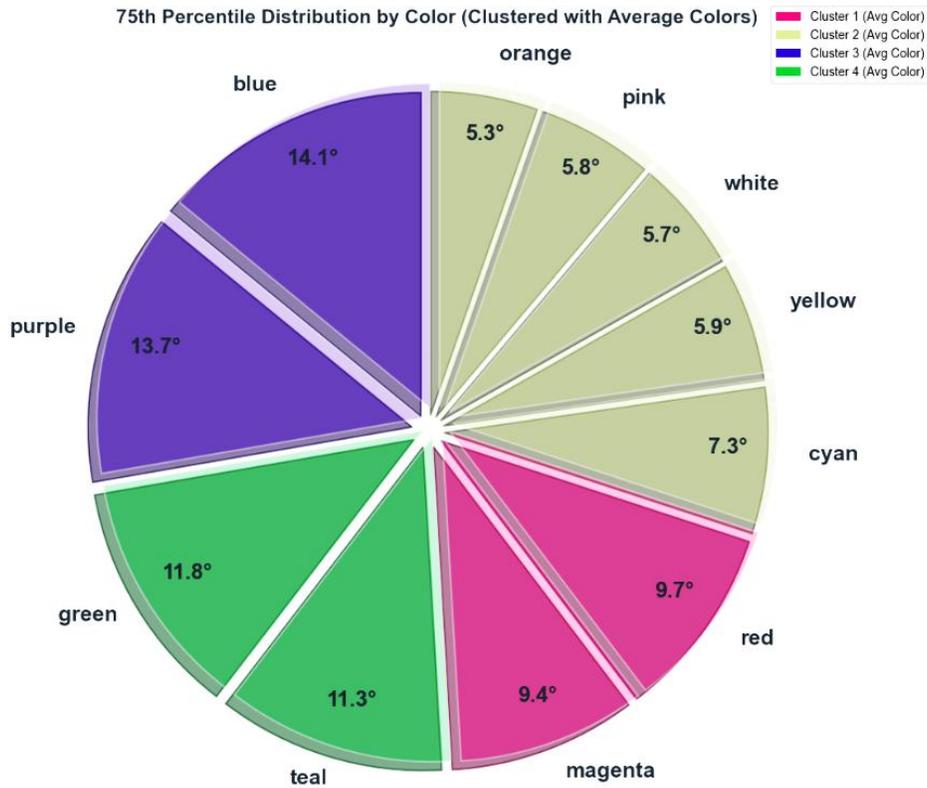

**Fig.11.** Clustered pie chart illustrating the distribution of angle prediction errors by color, based on their 75th percentile values. Colors are grouped using K-means clustering to highlight patterns in error magnitudes.

While using length and width maintains relatively stable angle errors across varying noise levels, incorporating color leads to increased errors as noise intensifies (Figure 12).

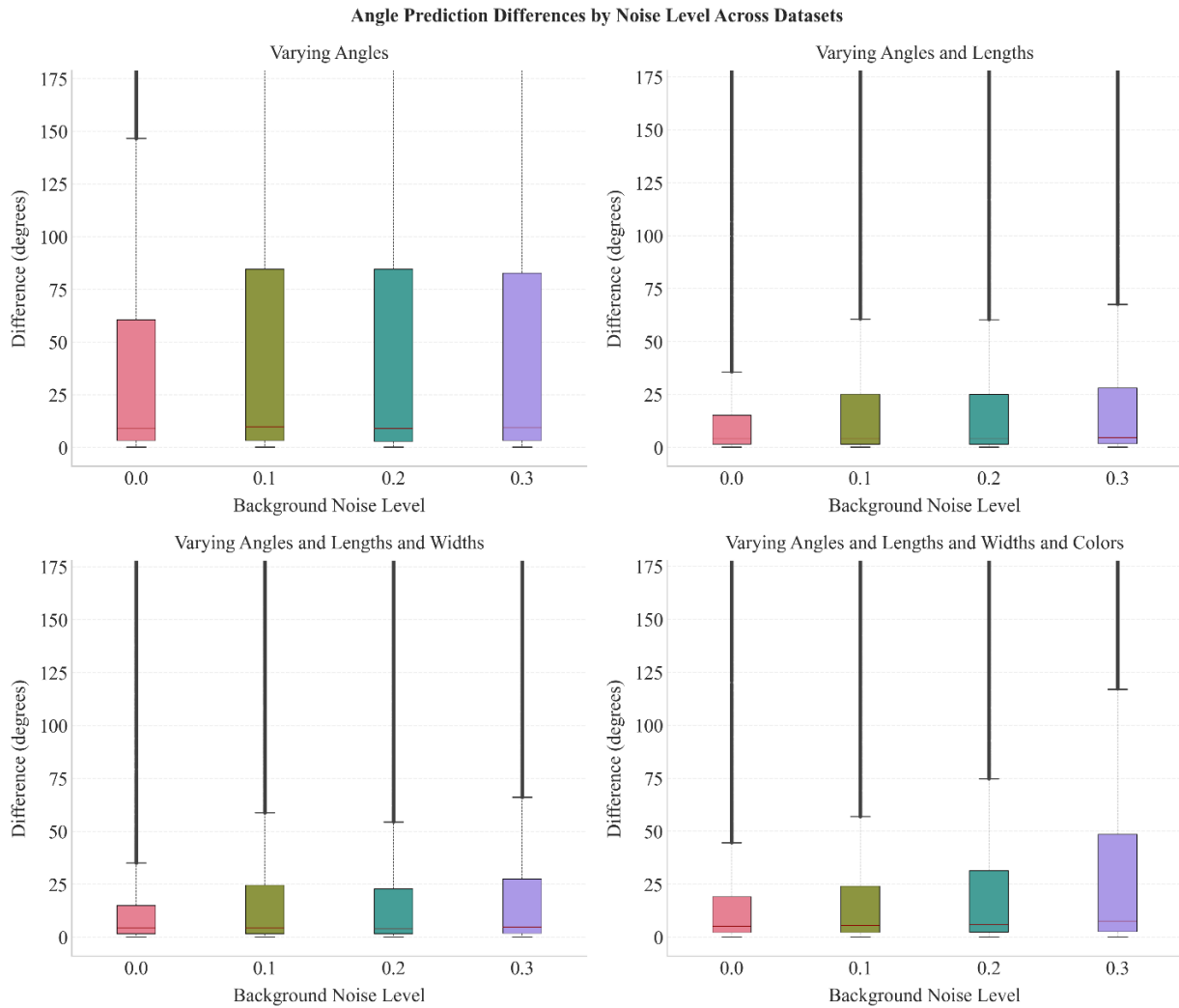

**Fig.12.** Boxplots showing angle prediction differences across varying background noise levels for four datasets with increasing visual complexity

The Figure 13 analyzes how different visual attributes—angle, length, width, and color—impact the 75th percentile error (in degrees) of angle predictions under increasing noise levels. Angle-only variations (Condition 'A') show the highest errors (60.55°–84.49°), indicating that predicting angles alone is inherently challenging, and noise has little effect on performance. Introducing length variations (Condition 'A+L') dramatically reduces errors (15.21°–27.99°), suggesting that length provides strong, noise-resistant cues for better angle estimation. Adding width (Condition 'A+L+W') has minimal additional impact, with errors nearly identical to those of length alone, implying width contributes little extra predictive value. However, incorporating color (Condition 'A+L+W+C') introduces noise sensitivity, as errors rise sharply with higher noise levels (19.08° to 48.35°), showing that color-based predictions degrade significantly under noise.

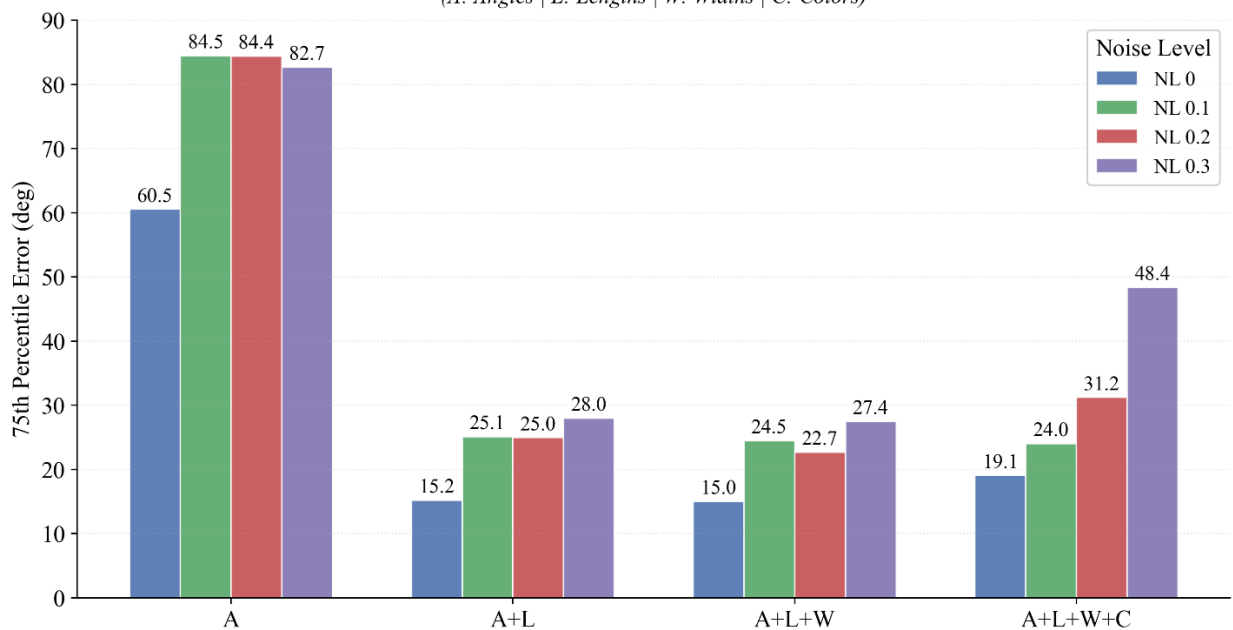

**Fig.13.** Noise Sensitivity of 75th Percentile Angle Error Across Visual Attributes—angle, length, width, and color: Angle-only (A) has high error (60.55°–84.49°); adding length (A+L) sharply reduces it (15.21°–27.99°). Width (A+L+W) has minimal effect, while color (A+L+W+C) increases noise sensitivity (19.08°–48.35°).

## 4. Discussion

The visual system exhibits biases in processing orientation, with neurons in the primary visual cortex (V1) showing preferential tuning for horizontal and vertical stimuli over oblique angles (Hubel et al., 1962). This anisotropy—termed the "oblique effect"—is reflected behaviorally in superior discrimination thresholds for cardinal orientations (Edwards et al., 1972). Mechanistically, V1 neurons tuned to horizontal orientations demonstrate narrower tuning widths, steeper response slopes, and stronger nonlinearities compared to those preferring other directions (Li et al., 2003). Such biases may arise from retinocortical projections or intracortical interactions, as horizontal-oriented stimuli evoke more robust responses in V1 than in higher areas like V2/V3 (Ge et al., 2020). The efficient coding hypothesis posits that these non-uniform sensitivities reflect statistical regularities in natural scenes, where cardinal orientations are overrepresented (Benjamin et al., 2022).

Parallel phenomena occur in color perception, where the brain groups continuous hue variations into discrete categories (e.g., red, green, blue). Clusters of color-preferring neurons ("globs") in the inferior temporal cortex encode these categories, with distinct populations responding to warm (red/yellow) versus cool (blue/green) hues (Bird et al., 2014). This categorical representation emerges mid-level in the visual hierarchy, optimizing task performance by compressing perceptually similar colors (Akbarinia, 2025). Notably, both orientation and color processing exhibit hierarchical refinement: early areas (e.g., V1) detect low-level features like edges, while higher areas integrate these into abstract representations (Fan et al., 2021). Like biological systems, deep neural networks prioritize learning statistically dominant features first (e.g., horizontal orientations or warm hues), with sensitivity patterns emerging from competitive interactions between subcircuits (Singh et al., 2024).

Another phenomenon observed in these systems is the role of phase transitions in shaping emergent properties. During training, deep neural networks undergo abrupt shifts in learning dynamics—marked by bumps or spikes in loss—as they acquire in-context learning abilities (Thilak et al., 2022). For instance, transformer models develop "induction heads" during a critical phase change, enabling pattern completion and abstract reasoning (Olsson et al., 2022). This suggest that neural systems—biological and artificial—leverage phase transitions to efficiently encode environmental statistics.

Orientation selectivity and color categorization may reflect optimal adaptations to natural scene regularities, while sudden shifts in network dynamics mirror the brain's developmental critical periods. Together, they highlight how complex representations arise from iterative, statistics-driven learning processes.

Vision Transformers (ViTs) have revolutionized computer vision, but their perceptual biases—particularly for orientation and color—remain underexplored compared to biological systems. Inspired by the brain's orientation selectivity and categorical color perception, we systematically investigated whether ViTs develop analogous biases through controlled experiments on synthetic datasets. We had generated four progressively complex datasets (varying angles → lengths → widths → colors) and fine-tuned a ViT with LoRA-adapted attention layers for multi-task regression. Our results revealed three key findings: (I) Orientation Selectivity: ViTs exhibited a robust bias toward horizontal lines (180°), demonstrated lower angle prediction errors for horizontal versus oblique orientations—mirroring the "oblique effect" observed in biological vision. This bias persisted across variations in line length, width, and color, suggesting fundamental anisotropy in ViT feature representations. (II) Color Sensitivity: Angle prediction errors varied categorically by color, with blue/purple hues inducing the highest errors and orange the lowest. K-means clustering grouped colors by error magnitude, revealing perceptual categories akin to human color perception. Notably, introducing color increased noise sensitivity, degrading angle prediction performance. (III) Phase Transitions in Learning: Training dynamics showed phase transitions (loss curve "bumps"), marking critical developmental milestones. Simpler datasets (angle-only) triggered earlier phase transitions, while complex datasets (angle+length+width+color) delayed them, reflecting modulated circuit formations.

This study has several limitations. First, the use of simple line stimuli may not fully capture perceptual biases in real-world scenarios, where complex geometries and noise distributions could interact differently with visual processing. Future work could address this by testing vision transformers (ViTs) on naturalistic datasets. Second, while phase transitions during training were identified, the specific ViT circuits underlying orientation and color biases remain unclear. A mechanistic dissection of these circuits—such as how early layers resemble V1 edge detectors versus deeper, more categorical representations—would deepen our understanding. Third, the ViT was trained on a regression task (angle prediction), leaving open whether biases generalize to classification or generative tasks (e.g., object detection or segmentation). Finally, the interaction between color and orientation biases was not rigorously explained; techniques like adversarial perturbations or gradient-based attribution could reveal conflicting feature representations. Future studies could explore how these biases propagate across layers and tasks.

**Appendix I: Synthetic Dataset Specifications**

### I.1.1 Dataset I: White Lines with Varied Angles

The image $I$ is a 3-channel $RGB$ image of size $W \times H \times 3$, where $W = 224$ and $H = 224$.

$$I \in \mathbb{R}^{W \times H \times 3}, \ I_{i,j,k} \in \{0, \ldots, 255\}$$

A line is defined by:

- A starting point $(x_1, y_1)$.
- An angle $\theta$ (in radians).
- A fixed length $L = 50$.

The endpoint $(x_2, y_2)$ is computed as:
$$x_2 = x_1 + L \cdot cos(\theta)$$
$$y_2 = y_1 + L \cdot sin(\theta)$$

where:
- $\theta \sim Uniform(0, 2\pi)$ (random angle).
- $x_1 \sim Uniform(L, W - L - 1)$ (random starting x-coordinate).
- $y_1 \sim Uniform(L, H - L - 1)$ (random starting y-coordinate).

The line is drawn with:
- Color: White $(255, 255, 255)$.
- Width: 2 pixels.

Gaussian noise is added to the image:
$$I_{noisy} = clip(I + N, 0, 255)$$

where:
- $N \sim N(0, \sigma^2)$, with $\sigma = noise\_level \times 255$.
- $Noise\ Level \in \{0, 0.1, 0.2, 0.3\}$ (Randomly chosen per image).
- $clip(\cdot)$ ensures pixel values remain in $[0, 255]$.

Each generated image is saved with metadata:
- Image ID: $Image\{i\}$ (sequential numbering).
- Angle: $\theta_{deg} = \theta \times \frac{180}{\pi}$ (converted to degrees).
- Start/End Points: $(x_1, y_1)$ and $(x_2, y_2)$.
- Noise Levels : $noise\_level$.
- Line Length: $L = 50$.

The dataset consists of $N = 50{,}000$ images, each with:
- A PNG file containing the noisy image with a white line.
- A metadata entry in an Excel file with the line parameters.

### I.1.2 Dataset II: White Lines with Varied Angles and Lengths

This dataset follows the same core structure as the first (black background, white lines, Gaussian noise), but introduces variable line lengths instead of fixed-length lines.

The line length $L$ is now randomly sampled (uniformly) between $[Lmin, Lmax]$:

$$L \sim Uniform\,(20, 100)$$

The starting point $(x_1, y_1)$ must ensure the entire line fits within the image:
$$x_1 \sim Uniform\,(L, W - L - 1)$$
$$y_1 \sim Uniform\,(L, H - L - 1)$$

The endpoint $(x_2, y_2)$ is computed as:
$$x_2 = x_1 + L \cdot cos(\theta)$$
$$y_2 = y_1 + L \cdot sin(\theta)$$

The metadata includes the fields: angle, start point, end point, noise level, and line length.

### I.1.3 Dataset III: White Lines with Varied Angles, Lengths, and Widths

This dataset extends the previous ones by introducing variable line widths in addition to variable lengths.

The line width w$w$ is now randomly sampled (discrete uniform) between $[w_{min}, w_{max}]$:
$$w \sim Uniform\{1, 2, 3, 4, 5\}$$

The start point $(x_1, y_1)$ must account for both line length L and width w$w$ to avoid clipping:
$$buffer = L + w$$
$$x_1 \sim Uniform(buffer, W - buffer - 1)$$
$$y_1 \sim Uniform(buffer, H - buffer - 1)$$

The endpoint $(x_2, y_2)$ is still computed as:
$$x_2 = x_1 + L \cdot cos(\theta)$$
$$y_2 = y_1 + L \cdot sin(\theta)$$

The metadata includes the fields: angle, start point, end point, noise level. Adds line length and line width.

### I.1.4 Dataset IV: Lines with Varied Angles, Lengths, Widths, and Colors

This dataset introduces variable line colors while maintaining variable lengths and widths same as Dataset III.

The line color $C$ is now randomly selected from a predefined set of $RGB$ tuples:
$$C \sim Uniform\{red, green, blue, \dots, teal\}$$

where each color maps to an $RGB$ triplet (e.g., $red = (255, 0, 0)$).

The metadata includes the fields: angle, start point, end point, noise level, line length, color RGB and color name.

Fig. I.1 illustrates the controlled variations in line properties (noise, width, length, angle, and color) across the synthetic datasets, enabling analysis of each parameter's impact on model performance.

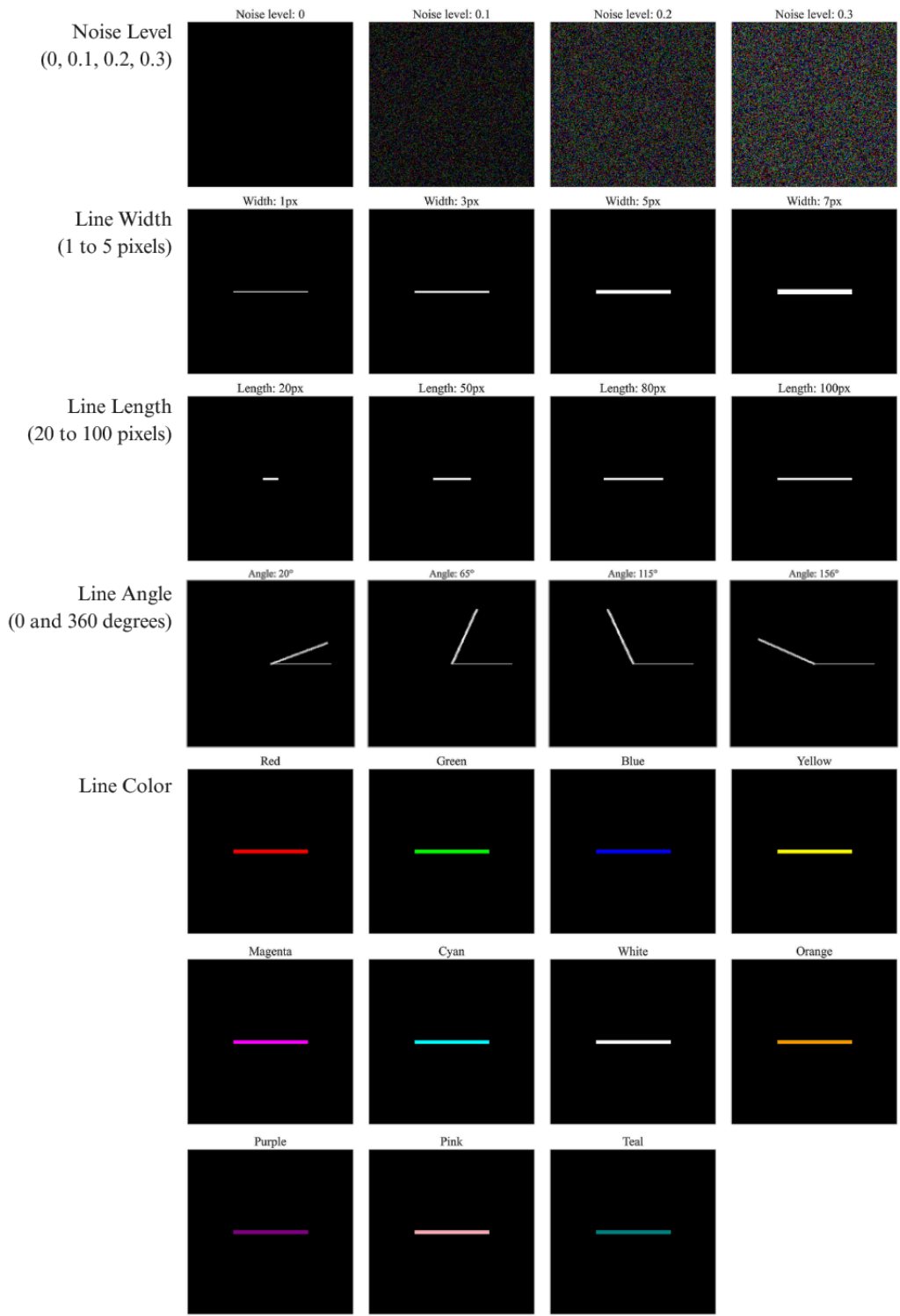

**Fig. I.1** Visualization of synthetic line image parameters including noise levels, widths, lengths, angles, and colors used for systematic vision transformer evaluation.